# Research on the Tender Leaf Identification and Mechanically Perceptible Plucking Finger for High-quality Green Tea

Wei Zhang[1, †]; Yong Chen[1, †]; Qianqian Wang[1]; Jun Chen[2, 3, *]

[1]College of Mechanical and Electronic Engineering, Nanjing Forestry University, Nanjing 210037, China

[2]Jiangsu Key Laboratory of 3D Printing Equipment and Manufacturing, Nanjing 210023, China

[3]College of Electrical and Automation Engineering, Nanjing Normal University, Nanjing 210023, China

[†]Wei Zhang and Yong Chen should be considered joint first author
[*]Correspondence: jun.chen@nnu.edu.cn

**Abstract:**

**BACKGROUND:** Intelligent identification and precise plucking are the keys to intelligent tea harvesting robots, which are of increasing significance nowadays. Aiming at plucking tender leaves for high-quality green tea producing, in this paper, a tender leaf identification algorithm and a mechanically perceptible plucking finger have been proposed.

**RESULTS:** Based on segmentation algorithm and color features, the tender leaf identification algorithm shows an average identification accuracy of over 92.8%. The mechanically perceptible plucking finger plucks tender leaves in a way that a human hand does so as to remain high quality of tea products. Though finite element analysis, we determine the ideal size of grippers and the location of strain gauge attachment on a gripper to enable the employment of feedback control of desired gripping force. Revealed from our experiments, the success rate of tender leaf plucking reaches 92.5%, demonstrating the effectiveness of our design.

**CONCLUSION:** The results show that the tender leaf identification algorithm and the mechanically perceptible plucking finger are effective for tender leaves identification and plucking, providing a foundation for the development of an intelligent tender leaf plucking robot.

## 1. Introduction

Tea is one of the three major beverages in the world, rich in nutritional and economic value[1,2]. However, tender leaves are still hand-plucked, which is time-consuming and laborious[3]. In recent decades, the increase of labor costs posts significant challenges in the harvesting of high-quality green tea, and thus constrains the sustainable development of the tea industry[4]. Two types of harvesters, hand-held and ride-on, have been developed as alternatives to manual labor, both of which harvest tea through shearing[5,6]. Although shearing is efficient for tea harvesting, it is not selective for tender and old leaves. Therefore, these harvesters are not helpful in producing high-quality green tea that preserves only tender leaves, as shown in Fig. 1 Schematic diagram of tender leaf. To develop automated tea harvesting machines that can efficiently pluck tender leaves of high-quality green tea, two key problems must be addressed: intelligent identification of tender leaves and high-speed biomimetic plucking.

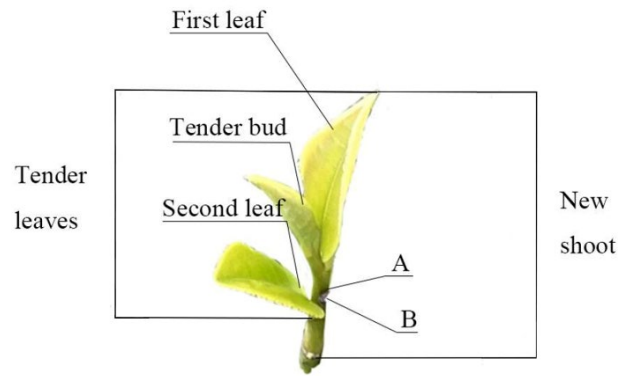

**Fig. 1** Schematic diagram of tender leaf

### 1.1 Tender Leaf Identification

In recent years, research on intelligent tea plucking has become a hot topic, and one important aspect is tender leaf identification, which is mainly based segmentation algorithm and deep learning. Zhang et al. proposed a method based on machine vision for real-time fresh tea leaves harvesting and received an identification accuracy of 90.3%[7]. Tang proposed a texture extraction method for green tea classification that combines non overlapping window local binary patterns and grayscale co-occurrence matrix[8]. Zhang et al. proposed a tea segmentation algorithm based on an improved watershed algorithm and obtained an identification accuracy of 95.79%[9]. Mukhopadhyay et al. proposed a new method for detecting tea diseases based on image processing technology[10].

Deep learning has shown excellent performance in identifying objects in complex backgrounds[11-14]. So far, there have been many studies using deep learning techniques in tender leaf identification, and convolutional neural networks such as YOLO[15-17] have been successfully applied to tender leaf identification. Yang et al. proposed a tender leaf identification method that combines semi supervised learning and image processing, and has identification accuracy of 92.62%[18]. Zhu et al. proposed an EfficientNet-B4-CBAM recognition method based on attention mechanism for Camellia oleifera identification[19]. Zhang et al. used two tea quality identification methods based on deep learning and transfer learning, and developed relevant application software, with identification accuracy of over 92.07%[20].

In this paper, segmentation algorithm is utilized for tender leaf identification due to its superior convenience and promising efficiency. Implementing segmentation algorithms allow practitioners to get rid of the complicated process of training deep networks through a large dataset on high-performance computing devices while requires only a small set of image samples to train on and low computing load. As a result, our method exhibits greater flexibility and generality, allowing the tea plucking machines to easily adapt to varies classes of tea trees without the need for heavy data training as a prerequisite.

## 1.2 Harvesting end effector

Fruit harvesting end effectors have been well-studied, with applications on harvesting apple[21,22], tomato[23,24], mushroom[25], kiwifruit[26], and strawberry[27]. Yet, tender leaf plucking is more challenging due to the fact that parts of the branch of tea trees are plucked instead of fruits, and that the color and morphology of the tender leaves are not as distinct as that of the fruits are from the branches. A few end effectors have been developed for branch plucking. Li et al. designed a tea plucking robot, used sleeve scheme as end effector and fixed the blade at the end of the sleeve to shear tender leaves[28]. Yang et al. used a Delta manipulator equipped with shear cutter to execute the trajectory tracking and tea plucking operations[29]. Zhu et al. designed a negative pressure guided end-effector for green tea harvesting, which solved the problem of visual positioning errors in tea tender leaves[30].

However, existing work on tea plucking end effectors utilizes a shear-based solution, i.e., sweeping the machine horizontally from the top of the tea tree and cut it indiscriminately to collect the cut. As a results, tender and old leaves are mixed and cannot be distinguished, producing only tea of low quality. To pluck tender leaves for producing high-quality green tea, we previously designed pneumatic and electric plucking fingers which mimics human hand plucking of tea[31,32]. The plucking fingers pluck tender leaves through pre-setting the stroke of plucking fingers. This paper designed a mechanically perceptible plucking finger to solve the problem that pre-set stroke cannot adapt the stems of different diameters. The mechanically perceptible plucking finger can real-time detect and close-loop control clamping force by clamping force measurement circuit and motor control circuit. The plucking finger can pluck tender leaves which meet the requirements of producing high-quality green tea and the success rate of plucking.

## 1.3 Contributions

This paper develops an autonomous mechanical plucking finger for high-quality green tea plucking. Our main contributions are threefold. Firstly, an intelligent tender leaf identification algorithm based on the segmentation algorithm is proposed, allowing tender leaves to be detected accurately. Secondly, a mechanically perceptible plucking finger is designed, which mimics human hand plucking of tea. Thirdly, a clamping force measurement circuit is attended to plucking finger control circuit, allowing plucking finger effectively pluck tender leaves with stems of different diameters.

## 2. Materials and methods

## 2.1 Tender leaf identification

### 2.1.1 Segmentation algorithm

Tender leaves can be distinguished from old leaves, soil, branches, etc., through color. For example, tender leaves are tender yellow while old leaves are dark green and the old stems are brown, shown in Fig. 2 Tea tree image. (a) Original tea tree image (b) Selected iteration area. (c) identification result.(a). Therefore, the $R$, $G$, and $B$ components of an image of tender leaves can be extracted in the color space using the color image segmentation algorithm, which identifies specified areas of colored images through finding regions with specific relationship of R, G and B components in these images.

It is assumed that linear superposition of RGB values of regions of interest in an image is greater than a segmentation threshold $T$ while the rest of the regions are lower than it, i.e.,

$$x \times c_r + y \times c_g + z \times c_b \geq T, \qquad (1)$$

where $x$, $y$, and $z$ represent the segmentation coefficients of the $c_r$, $c_g$, and $c_b$ components, respectively. In this way, the region of interest can be accurately extracted from the image by establishing a relationship between the RGBs to find the difference between the region of interest and other regions.

To determine $x$, $y$, $z$ and $T$, a series of samples of pixels must be collected from the images. An iteration method is then applied to these samples, which iterates the values of each parameter in a certain range by some step length for each pixel until a combination of the parameters is found, if exists, such that a desired percentage of pixels satisfy Equation 1. Otherwise, one may reset the ranges of parameters and step length for more iteration attempts.

### 2.1.2 Parameter fitting

To identify the tender leaves, we took ten images of tea tree of different varieties, from different scenarios and growth stages. From each image, ten 10×10-pixel rectangular boxes were sampled at regions of different parts of tender leaves, an example of which is shown in red boxes in Fig. 2 Tea tree image. (a) Original tea tree image (b) Selected iteration area. (c) identification result.. Meanwhile, another ten 10×10-pixel rectangular boxes were sampled at other regions including old leaves, old branches, soil, shadows, etc., as shown in blue boxes in Fig. 2 Tea tree image. (a) Original tea tree image (b) Selected iteration area. (c) identification result.(b). Therefore, a total number of 20,000 pixels, i.e., RGB values, were sampled for segmentation algorithm parameter fitting using the iteration method.

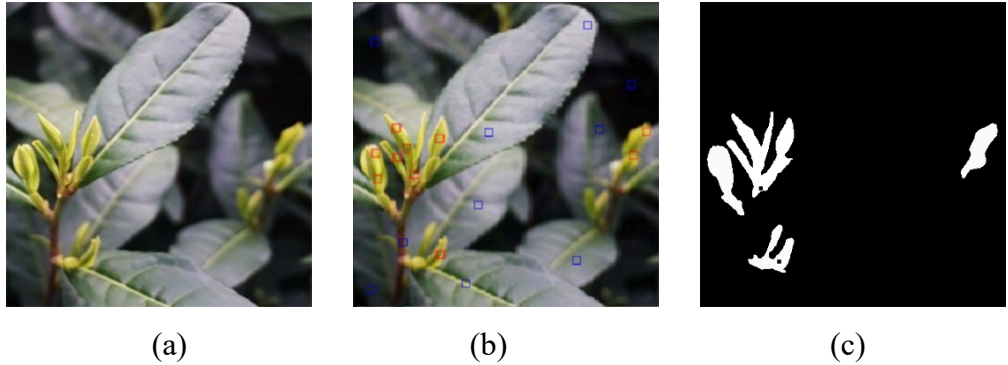

(a)          (b)          (c)

**Fig. 2** Tea tree image. (a) Original tea tree image (b) Selected iteration area. (c) identification result.

Referring to excess green index $Exg = 2c_g - c_r - c_b$[33], the segmentation coefficients $x$, $y$, and $z$ are set as [-3, 3], and the segmentation threshold $T$ is set as [0,255]. The iteration step size of the segmentation coefficients $x$, $y$, and $z$ were set as 0.05, and that of the segmentation threshold $T$ was set as 0.5. To trade-off between identification accuracy and computational efficiency based on our previous experience, the iteration process terminates when a set of parameters is found such that at least 98% of the tender leaf pixels meet Equation 1, and that no more than 2% of the background pixels meet Equation 1 at the same time, i.e.,

$$\begin{cases} n > 0.98 \times N, \\ m < 0.02 \times M, \end{cases} \qquad (2)$$

where $n$ and $m$ represent the numbers of tender leaf and background pixels that meet Equation

1, respectively, and $N$ and $M$ represent the total numbers of tender leaf and background pixels that meet Equation 1, respectively. Results are recorded in Table 1 for each image, where $x, y, z$ and $T$ were fit to 0.764, 0.392, -1.157, and 90.3, respectively via averaging the ten images.

**Table 1.** The iteration results of ten images

|  | $x$ | $y$ | $z$ | $T$ |
| --- | --- | --- | --- | --- |
| 1 | 1.62 | -0.31 | -1.73 | 74 |
| 2 | 0.81 | 0.61 | -1.17 | 126 |
| 3 | 0.38 | 0.48 | -1.04 | 82 |
| 4 | 0.48 | 0.66 | -1.18 | 72 |
| 5 | 0.76 | 0.40 | -0.99 | 88 |
| 6 | 0.89 | 0.52 | -1.05 | 132 |
| 7 | 0.74 | 0.49 | -1.04 | 94 |
| 8 | 0.67 | 0.44 | -1.41 | 64 |
| 9 | 1.03 | 0.05 | -0.92 | 92 |
| 10 | 0.26 | 0.58 | -1.04 | 79 |
| average | 0.764 | 0.392 | -1.157 | 90.3 |

To extract tender leaf regions, the images were then binarized. For each pixel at location $q$ with $c_r$, $c_g$, and $c_b$ components, the binarized value is defined by

$$f(q) = \begin{cases} 255, & 0.764 \times c_r + 0.392 \times c_g - 1.157 \times c_b \geq 90.3 \\ 0, & others. \end{cases} \quad (4)$$

An example result is shown in Fig. 2 Tea tree image. (a) Original tea tree image (b) Selected iteration area. (c) identification result.(c).

## 2.2. Design of mechanically perceptible plucking finger

### 2.2.1 Overall structure of mechanically perceptible plucking finger

Similar to our previous work[31], the main structure of our designed mechanically perceptible plucking finger consists of two pairs of grippers, four sleeves and three stepper motors, demonstrated in Fig. 3 Structure diagram of mechanically perceptible plucking finger. The end effector works as follows:

(1) The upper gripper is connected to the mandrel by a clamping arm and two hinges, and to the inner sleeve through a hinge. The point-A stepping motor for switching rotates to make the inner sleeve move upward and downward, which leads the upper gripper to open and close.

(2) The lower gripper is connected to the outer sleeve by a clamping arm and two hinges, and to the middle sleeve through a hinge. The point-B stepping motor for switching rotates to make the outer sleeve move upward and downward, which leads the lower gripper to open and close.

(3) The rotation of the point-A step motor for pulling makes the mandrel move up and down, thus driving the upper gripper both to move up and down and to open and close simultaneously. Therefore, in order to keep the upper gripper moving upward with the clamping action, it is necessary to set the point-A stepping motor for switching and the point-A step motor for pulling to rotate with the same angular velocity. When the upper and lower grippers have fixed a tender leaf, the point-B stepping motor for switching stops rotating, and the point-A stepping motor for switching rotates to drive the inner sleeve upward. Meanwhile, the point-A step motor for pulling rotates at the same speed to drive the mandrel upward while maintaining relatively stationary between the mandrel and the inner sleeve. Therefore, the upper gripper clamps the point-A of the tender leaf in Fig. 1 Schematic diagram of tender leaf and move upward to achieve the separation of the point-A and the point-B, to complete the bionic harvesting of it.

**Remark 1:** Our design allows tender leaves to be plucked by pulling the stems of tea trees rather than by shearing. In this way, it avoids oxidizing and blackening of the cut sections of the tender leaves when exposed to the air, which in turn improves the quality of the tea products.

**Remark 2:** The grippers are attached to the body of the plucking finger so that when two pairs of grippers are close together, they are at the same angle to the mandrel. Therefore, the stresses and strains in the two pairs of grippers are the same when clamping the tea tree stems.

**2.2.2 Mechanically perceptible mechanism to adapt stem diameters**

To effectively pluck tender leaves, appropriate clamping force must be applied by the plucking finger to ensure that tea stems are firmly clamped without being crushed and damaged. Our previous design achieves this through predefining specific travel distances for the grippers in which way that the grippers can fit a constant diameter of the tea stem[31]. However, desired performances are not obtained in the cases that tea stems have different diameters.

Therefore, a mechanically perceptible plucking finger is proposed. In this way, when the two pairs of grippers move close to one another to clamp a tea stem, it stops once an appropriate clamping force is reached. This is achieved by 1) designing two pairs of grippers that can produce slight deformation during clamping, 2) attaching a resistance strain gauge to the deformable part of the grippers, and 3) adding a tender leaf gatherer at the end of each gripper to gather tender leaves for improved success rate of stem clamping. Therefore, we design a novel structure of grippers and choose to utilize aluminum alloy grippers with smaller rectangular cross-sectional dimensions, making it easier to measure the clamping force based on gripper deformation.

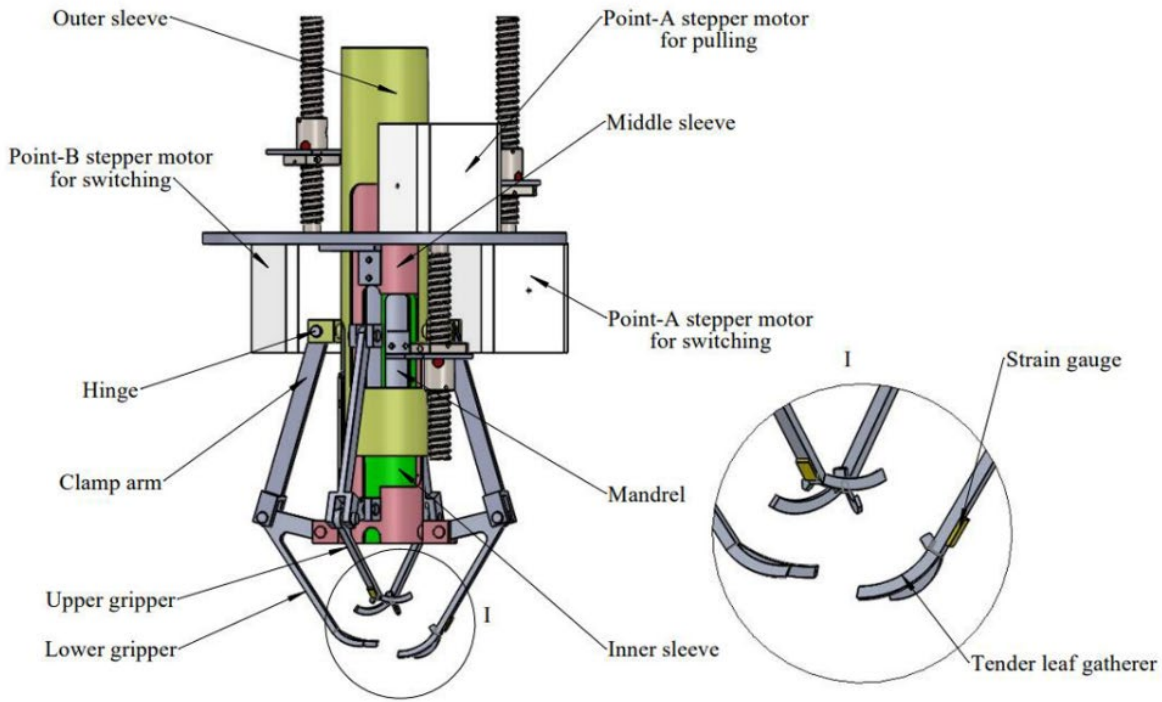

**Fig. 3** Structure diagram of mechanically perceptible plucking finger

To determine cross-sectional dimensions of grippers and strain gauge attachment locations, finite element analysis was performed using ANSYS to find out the degree of deformation of the grippers and the most bendable part of them. According to Remark 2, finite element analysis was only performed on the upper gripper.

Preliminary results have shown that the clamping force applied to the tender leaves should be between 3N and 4N in order to prevent slipping or breaking of the stem when plucking the tender leaves[31]. Therefore, considering the clamping and pulling forces of tender leaves, the clamping force of the grippers is set to 4N. Through force analysis of the gripper, it can be obtained that $F_1$ is 3.57N and $F_2$ is 2.41N (Fig. 4). $F_1$ and $F_2$ are added to the gripper, and fixing force is added at the connection between the gripper and the plucking finger. The imported gripper model was meshed and the size of mesh division was specified as 1mm. Considering that the selected strain gauge has a width of 3mm, and through continuous modification of the gripper parameters, the cross-sectional parameter of the gripper is determined to be 3×3mm. The stress distribution and deformation distribution of the gripper were obtained from finite element analysis (Fig. 5).

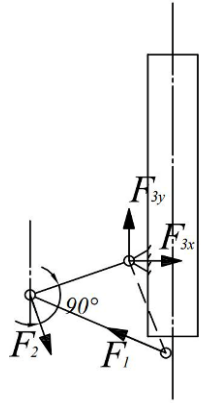

**Fig. 4** Force analysis

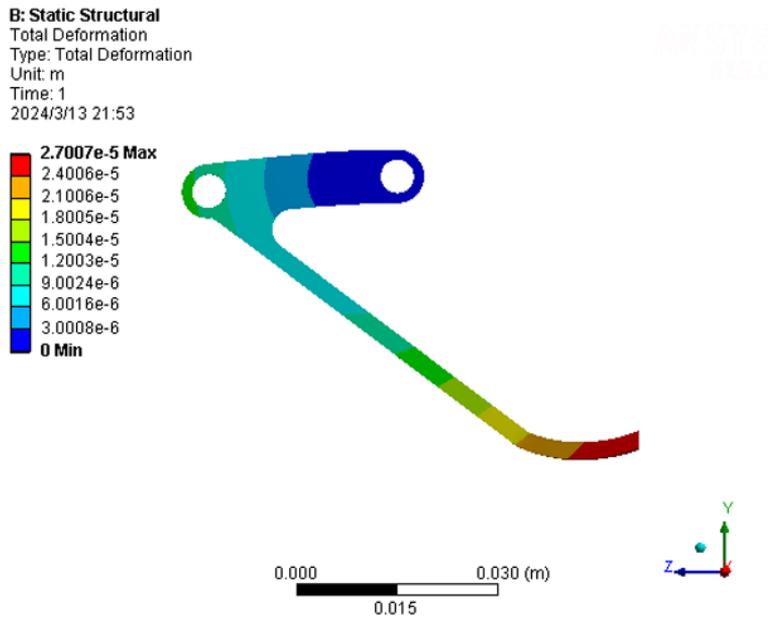

**Fig. 5** Distribution map of gripper deformation

From distribution map of upper deformation, the maximum deformation of the gripper is 0.027mm, located at the clamping point of the upper gripper (Fig. 5). Strain gauge can detect deformation at the micrometer level. By comparing with the results, the deformation generated by clamping force during clamping can be easily detected by strain gauge. Based on modified parameters and selected materials, a mechanically perceptible plucking finger was made (Fig. 10).

**2.2.3. Control system design**

**Circuit:** The control system circuit of the plucking finger uses STM32 as the main controller, mainly consists of clamping force measurement circuit and stepper motor control circuit. Since the upper and lower grippers of the plucking finger were subjected to the same force according to Remark 2, this paper only presents the design of the upper grippers clamping force measurement circuit. The clamping force measurement circuit uses a foil strain gauge to build an electric bridge, demonstrated in Fig. 6. Strain gauges are attached to the maximum deformation point of the upper clamp, connecting the SG1+ and SG1- of the bridge circuit. When a tea stem is clamped, causing the grippers to deform, the resistance value of the strain gauge will change and the electric bridge circuit outputs a voltage variation. After completing the digital-to-analog conversion through STM32, the control system obtains the clamping force of the upper gripper on the stem. According to the clamping force measured by the clamping force measurement circuit, the stepper motor control circuit controls the three stepper motors to achieve the opening, closing and pulling actions of the upper and lower grippers.

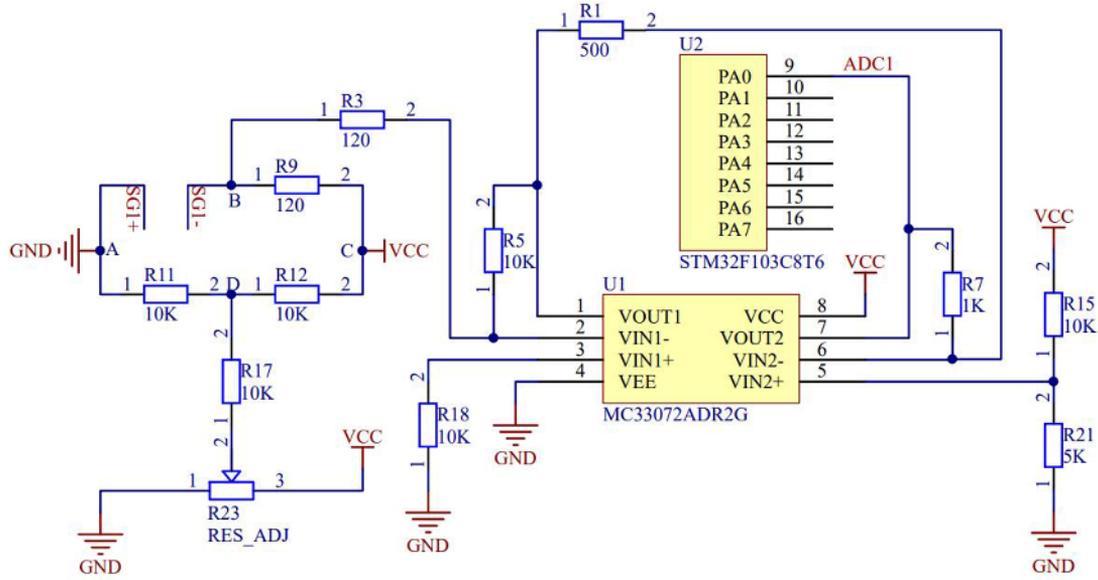

**Fig. 6** Clamping force measurement circuit

**Close loop control law:** As the plucking finger starts to pluck, the upper and lower grippers close in on each other to clamp the stem. When the clamping force reaches 4N, the lower grippers are kept still. While the clamping angle of the upper grippers are kept unchanged, the upper grippers pull the tender leaves upwards, causing the tender leaves broken between the two clamping points, thereby achieving the separation of the tender leaves from the stem. After completing the separation of tender leaf, the upper and lower grippers open outward. After detecting a clamping force of 0N, all mechanisms were reset and ready to pluck next tender leaf.

## 3. Results and discussion

### 3.1. Identification results of different tender leaf varieties

To evaluate the efficacy of tender leaf identification, we define the identification accuracy $R_i$ the percentage of the number of identified tender leaves $n$ to the actual number of tender leaves $N$, i.e.,

$$R_i = \frac{n}{N} \times 100\%. \tag{5}$$

To evaluate the efficacy of tender leaf identification, we define the identification accuracy $R_i$ the percentage of the number of identified tender leaves $n$ to the actual number of tender leaves $N$, i.e.,

$$R_m = \frac{m}{N} \times 100\%. \tag{6}$$

To examine the generality of our proposed identification algorithm for different tender leaf varieties, a total of fifty images of tender leaves from Jiukeng population species, Longjing Longye, Shuchazao, Longjing 43, and Chinese tea 108 were tested. Qualitative results are shown in Fig. 7, where Fig. 7(a)-(d) show the original images of different tender leaf varieties, and Fig. 7(e)-(f) show the identification results. The identification results of average $R_i$ and $R_m$ are shown in Table 2. Our identification algorithm shows the highest success rates for Chinese tea 108, with identification accuracy of 98.98%, and the lowest for Jiukeng population, with identification accuracy of 92.12%. In general, the algorithm has strong versatility and generality, with an average identification accuracy of 96.67% on varies kinds of tender leaves.

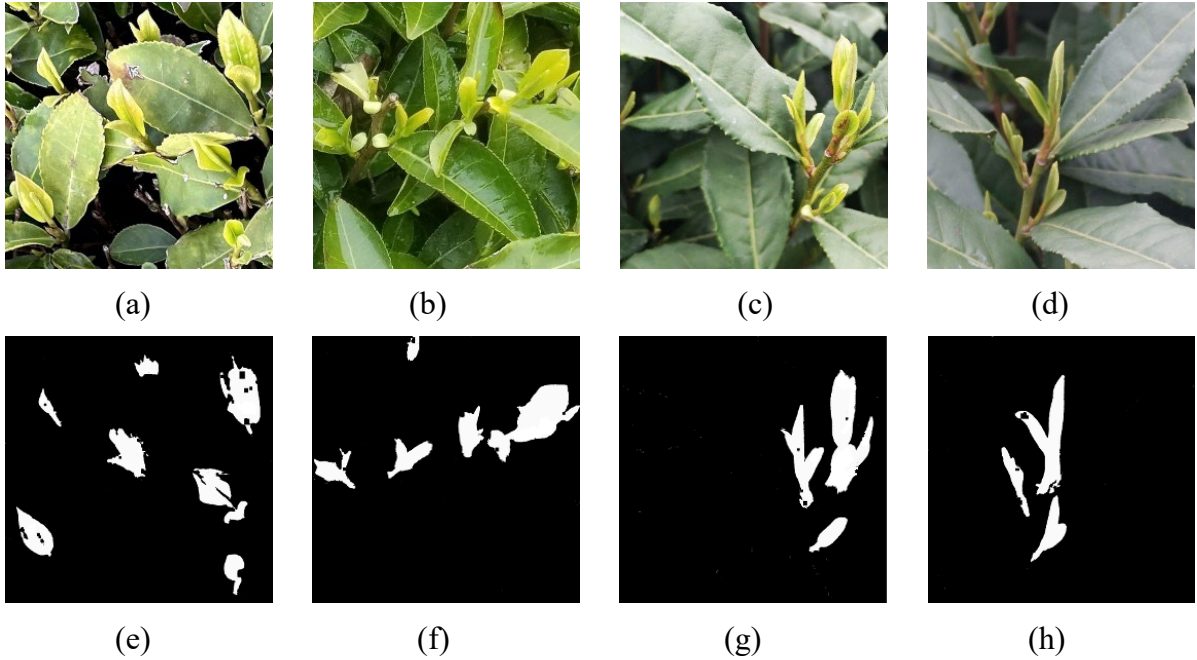

(a)　　　　　　　(b)　　　　　　　(c)　　　　　　　(d)

(e)　　　　　　　(f)　　　　　　　(g)　　　　　　　(h)

**Fig. 7** Identification results of different tender leaf varieties. **(a)** Original image of Jiukeng population species. **(b)** Original image of Longjing Longye. **(c)** Original image of Longjing 43. **(d)** Original image of Chinese tea 108. **(e)** Identification result of Jiukeng population species. **(f)** Identification result of Longjing Longye. **(g)** Identification result of Longjing 43. **(h)** Identification result of Chinese tea 108.

**Table 2.** Comparison of the identification results of different tender leaf varieties

| Tender leaf variety | $R_i$ | $R_m$ |
| --- | --- | --- |
| Jiukeng population | 92.12% | 7.36% |
| Longjing Longye | 96.94% | 3.36% |
| Longjing 43 | 98.64% | 1.28% |
| Chinese tea 108 | 98.98% | 1.16% |
| Average | 96.67% | 3.29% |

### 3.2. Identification results of tender leaves under different lighting conditions

To examine the effect of light factor on the tender leaf identification algorithm, we tested a total of twenty images of tea trees under strong light irradiation and shading environment. Qualitative identification results are shown in Fig. 8 where Fig. 8(a)-(b) show the original images of different light conditions, and Fig. 8(c)-(d) show the identification results. The identification results are shown in Table 3, where we can see that strong light reduces the success rates of tender leaf identification. In general, the algorithm can effectively identify tender leaves at different light conditions, with average identification accuracy of 92.8%.

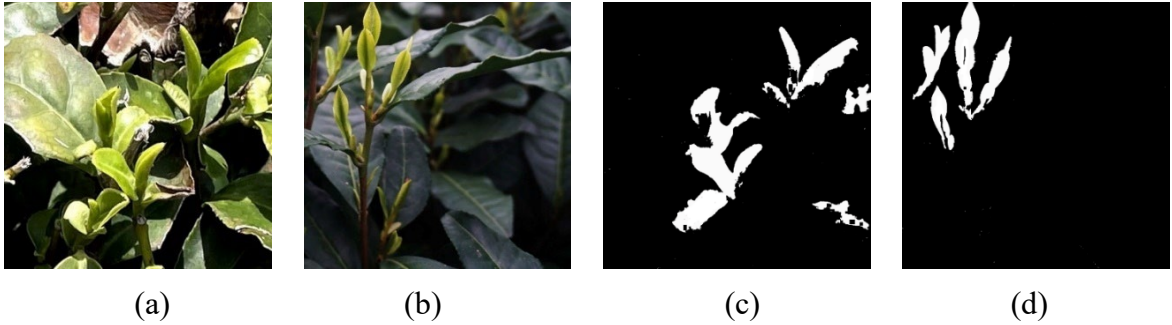

(a)          (b)          (c)          (d)

**Fig. 8** Identification results of tender leaf under different light conditions. **(a)** Original image of tender leaf under strong light irradiation. **(b)** Original image of tender leaf under shade environment. **(c)** Identification result of tender leaf under strong light irradiation. **(d)** Identification result of tender leaf under shade environment.

**Table 3.** Comparison of tender leaf identification results under different light conditions

| Light condition | $R_i$ | $R_m$ |
| --- | --- | --- |
| Strong light irradiation | 87.52% | 13.76% |
| Shadow occlusion | 98.08% | 1.92% |
| Average | 92.8% | 7.84% |

### 3.3. Identification results of tender leaf at different growth stages

To examine the generality of our tender leaf identification algorithm for tea at different growth stages, a total of twenty tea images from mid-March and mid-April were tested. Qualitative identification results are shown in Fig. 9 where Fig. 9(a)-(b) show the original images of different growth stages, and Fig. 9(c)-(d) show the identification results. The identification results are shown in Table 4. In general, the algorithm has a good identification effect on early-spring tender leaf, with average identification accuracy of 97.52%

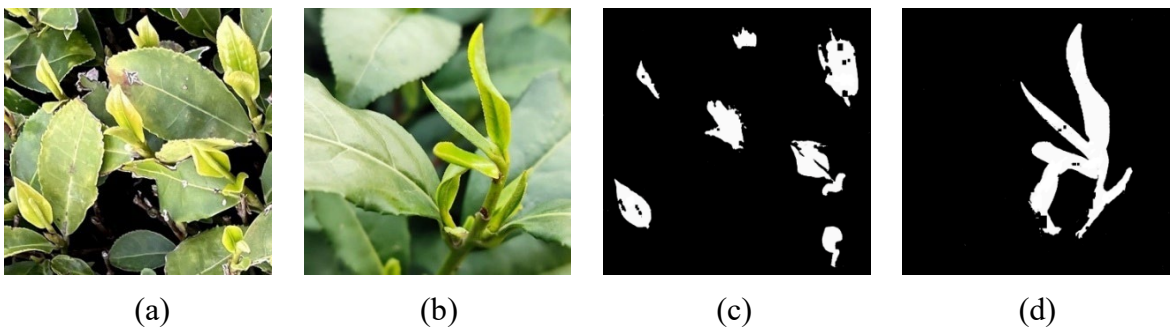

(a)          (b)          (c)          (d)

**Fig. 9** Identification results of tender leaves at different growth stages. **(a)** Original image of tender leaf in mid-March. **(b)** Original image of tender leaf in mid-April. **(c)** Identification result of tender leaf in mid-March. **(d)** Identification result of tender leaf in mid-April.

**Table 4.** Comparison of tender leaf identification results at different growth stages

| Growth stage | $R_i$ | $R_m$ |
|---|---|---|
| mid-March | 96.32% | 4.38% |
| mid-April | 98.72% | 1.36% |
| average | 97.52% | 2.87% |

### 3.4. Plucking results

An electric control system based on a mechanically perceptible plucking finger prototype was built as Fig. 10. The electric control system mainly consists of a control circuit board, a DC power supply, three stepper motor drivers, three stepper motors, and a mechanically perceptible plucking finger.

This experiment used tender leaves of Yuhua tea as the experimental material. The Yuhua tender leaves were plucked twice at Zhongshan Ling Tea Plantation in Nanjing in mid-April 2023, with a quantity of 30 and 50. After plucking, they were immediately returned to the laboratory for tender leaves plucking experiment. The plucking process is shown in Fig. 11. The mechanically perceptible plucking finger effectively achieves the separation of tender leaves and stems (Fig. 11). The statistical results of the plucking experiment are shown in Table 5. The results showed that the mechanically perceptible plucking finger has a success rate of 92.5% for plucking tender leaves.

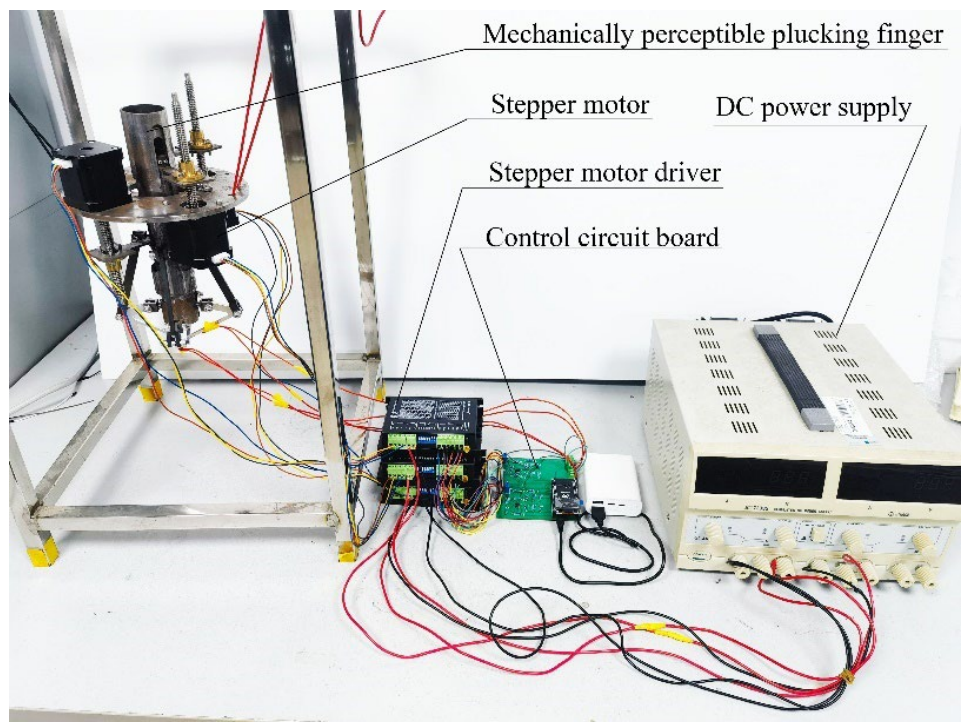

**Fig. 10** Electric control system for flexible plucking finger

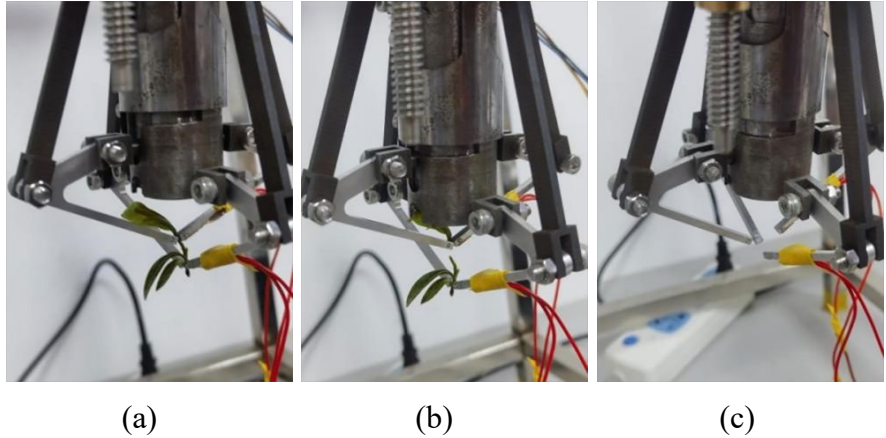

| (a) | (b) | (c) |

**Fig. 11** Plucking process. **(a)** Clamping tender leaves. **(b)** Tender leaves separation. **(c)** Plucking completed.

**Table 5.** Results of plucking experiments

|  | sample size | Number of clips broken | Slipping number | Incomplete number | Number of successful harvests | Plucking success rate |
| --- | --- | --- | --- | --- | --- | --- |
| First experiment | 30 | 2 | 0 | 1 | 27 | 90% |
| Second experiment | 50 | 2 | 1 | 0 | 47 | 94% |
| total | 80 | 4 | 1 | 1 | 74 | 92.5% |

In the plucking experiment, there are six samples which failed to pluck. The reason for plucking failure is that the clamping force measurement circuit measured the inaccurate clamping force or the control system failed to timely control the motors start and stop. The underlying reason is that there are tolerances in processing and long-term use of parts, which cause the plucking finger to be misaligned during plucking, resulting in incomplete tender leaves or slipping on the stems.

## 4. Conclusion

Since the key techniques of intelligent tea harvest are tender leaf identification and plucking, a new tender leaf identification algorithm was proposed based on color features and segmentation method. The algorithm has strong versatility for the identification of tender leaves of different varieties, under different lighting conditions and at different growth stages, with an average identification accuracy of over 92.8%. A mechanically perceptible plucking finger which mimics the manual plucking action was innovatively proposed. The mechanically perceptible plucking finger which can pluck different diameters of tender leaves and stems meets the requirement of tender leaves for high-quality green tea producing. The success rate of tender leaf plucking reaches 92.5%. In the future, the identification accuracy and plucking success rate will be further improved, tender leaf plucking robot which combine intelligent and mechanically perceptible plucking finger will be manufactured, and the plucking experiment will be carried out in tea plantation.

ACKNOWLEDGEMENTS

This research was supported by the National Natural Science Foundation of China (Grant No. 32072498).

CONFLICT OF INTEREST STATEMENT

The authors declare no conflict of interest.